\documentclass{article}
\usepackage{spconf,amsmath,graphicx,amssymb}

\title{Learning Modality-Invariant Representations \\ for Speech and Images}
\name{Kenneth Leidal, David Harwath, James Glass}
\address{
  Computer Science and Artificial Intelligence Laboratory \\
  Massachusetts Institute of Technology \\
  Cambridge, MA 02139, USA \\
  \texttt{\{kkleidal, dharwath, glass\}@mit.edu}}
  
\newcommand{\inp}[2]{x_#1^{(#2)}}
\newcommand{\enc}[2]{f_#1(\inp{#1}{#2})}
\newcommand{\img}{v}
\newcommand{\aud}{a}
\newcommand{\repdim}{z}
\newcommand{\imgspace}{\mathcal{V}}
\newcommand{\capspace}{\mathcal{A}}
\newcommand{\igloss}{\mathcal{L}_\text{IG}}
\newcommand{\simloss}{\mathcal{L}_\text{Sim}}
\newcommand{\wdloss}{\mathcal{L}_\text{WD}}

\newcommand{\lamig}{\lambda_\text{IG}}
\newcommand{\lamwd}{\lambda_\text{WD}}
\newcommand{\posterior}[2]{\mathcal{N} \left( \mu_{#1}(\inp{#1}{#2}), \text{diag} \left( \sigma^2_{#1}(\inp{#1}{#2}) \right) \right)}
\newcommand{\post}[2]{\hat{p}_{f_{#1}\ |\ \inp{#1}{#2}}}

\usepackage{float}
\newfloat{twocolequfloat}{t}{zzz}
\floatname{twocolequfloat}{Equation} 

\usepackage{tikz}
\usetikzlibrary{shapes.geometric, arrows}
\usetikzlibrary{positioning}
\tikzstyle{input} = [rectangle, rounded corners, minimum width=2cm, minimum height=1cm,text centered, draw=black, fill=red!30]
\tikzstyle{latent} = [rectangle, rounded corners, minimum width=2cm, minimum height=1cm,text centered, draw=black, fill=green!30]
\tikzstyle{output} = [rectangle, rounded corners, minimum width=2cm, minimum height=1cm,text centered, draw=black, fill=blue!30]
\tikzstyle{hidden} = [rectangle, minimum width=2cm, minimum height=1cm, text centered, draw=black, fill=orange!30]
\tikzstyle{arrow} = [thick,->,>=stealth]
\tikzstyle{line} = [thick,-,>=stealth]

\usepackage{enumitem}

\begin{document}
%
\maketitle

\begin{abstract}
In this paper, we explore the unsupervised learning of a semantic
	embedding space for co-occurring sensory inputs.
Specifically, we focus on the task of learning a semantic 
	vector space for both spoken and handwritten digits using the
	TIDIGITs and MNIST datasets.
Current techniques encode image and audio/textual inputs
	directly to semantic embeddings.
In contrast, our technique maps an input to the mean and log variance vectors of
	a diagonal Gaussian from which sample semantic embeddings are drawn.
In addition to encouraging semantic similarity between co-occurring inputs,
	our loss function includes a regularization term borrowed from
	variational autoencoders (VAEs) which 
    drives the posterior distributions over embeddings to be unit Gaussian.
We can use this regularization term to filter out modality information
	while preserving semantic information.
We speculate this technique may be more broadly applicable
	to other areas of cross-modality/domain information retrieval and
    transfer learning.
    

\end{abstract}
\section{Introduction}

The high-level goal of this paper is to train a neural model to learn a
	semantic embedding space into which speech audio and image inputs can be mapped.
This goal is inspired by the fact that children are able to learn through associating stimuli during
	their early years.
For example, a child hearing his or her mother pronounce ``seven'' or write a ``7''
	might learn to think of the same concept upon hearing or seeing either.
In fact, Man et al. showed that the temporoparietal cortex of the human brain produces content-specific
    and modality-invariant neural responses to audio and visual stimuli \cite{man2012sight}.

In our method, we map image and audio inputs to the parameterizations of diagonal Gaussians
	representing the posterior distribution over semantic embeddings.
We then sample embeddings from this distribution and use a loss function which encourages
	samples from paired audio
	and image inputs to be more similar than mismatched pairs of audio and images.
Although this objective has been shown to encourage the embedding space to be rich semantically \cite{harwath16},
    in this paper we explore methods of better encouraging modality-invariance.
That is, not only should semantically relevant content be clustered in the embedding space,
    but the distributions of embeddings for semantically equivalent audio and images
    should be the same.
This goal is based on the assumption that information concerning modality is
	noise for tasks requiring only the semantic content of the sensory input.
  
To drive the posterior distributions over embeddings to be the same for semantically equivalent
	inputs across modalities, we introduce a term to the objective which regularizes the amount of
    information encoded in the semantic embedding.
The term, borrowed from variational autoencoders (VAEs), is the sum of the KL divergences of
	the posterior distributions from the unit Gaussian.
Our results suggest that when this regularization term is
	increased from zero during hyperparameter tuning, modality-information
    tends to be filtered out prior to semantic-information.
We believe this technique has the potential to be useful for
	modality-invariant and domain-invariant applications.

\section{Previous Work}

\begin{twocolequfloat*}
\hfill
\begin{equation}
\label{eq:lsim}
\begin{split}
\simloss =
\sum_{\substack{i : 1 \leq i \leq N \\ j \sim \text{Uniform}(\left\lbrace x : 1 \leq x \leq N \wedge x \neq i \right\rbrace)}}
&\max \left(0, 1 - \text{sim}\left( \enc{\aud}{i}, \enc{\img}{i} \right) + \text{sim}\left( \enc{\aud}{i}, \enc{\img}{j} \right) \right) \\
+ &\max \left(0, 1 - \text{sim}\left( \enc{\aud}{i}, \enc{\img}{i} \right) + \text{sim}\left( \enc{\img}{i}, \enc{\aud}{j} \right) \right)
\end{split}
\end{equation}
\end{twocolequfloat*}
The unsupervised problem of learning semantic relations through the co-occurrence
	and lack of co-occurrence of sensory inputs is an increasingly attractive
    pursuit for researchers \cite{harwath16,wang2016comprehensive,saito2016demian,aytar2016soundnet}.
This attraction is primarily due to the expense of attaining labels for data.
The ability
	to learn semantic relevance with input pairings alone unlocks the potential
    of training models using inexpensively-collected
    data with the only supervisory signal being the co-occurrence of
    sensory inputs \cite{wang2016comprehensive}. 
In addition, the learned semantic space has
	direct practical applications.
One particular application of a semantic space is cross-modality transfer learning:
	using paired inputs from two
	modalities and labels for one modality to learn how to predict labels for the unlabeled
    modality.
Aytar, Vondrick, and Torralba \cite{aytar2016soundnet} use a teacher-student model on videos
	to transfer knowledge from
	pretrained ImageNet and Places convolutional neural networks (CNNs) identifying object and scene information
	in images to train a CNN run on the raw audio waveform from the video to
    recognize the same information.
In Aytar et al.'s model,
	the shared semantic space consists of the two distributions over
	objects and scenes as opposed to being an arbitrary (yet highly linearly
    correlated) semantic space, as is the case in our model.
Wang et al. \cite{wang2016comprehensive} gave a comprehensive overview of existing approaches to another
	practical application of shared semantic spaces: cross-modality information
	retrieval.
The task is formulated as follows:  given an input of one modality, find related instances of another modality.
In 2016, Harwath et al.~\cite{harwath16} presented a method to learn a semantic embedding space into which
	images and spoken audio recordings of captions of the images
    could be mapped.
They evaluated their method by looking at the cross-modality retrieval recall scores: e.g., given
	an image and $N$ audio captions, which of the $N$ audio captions describes the
    image?
K. Saito et al.~developed an adversarial neural architecture
	to learn modality-invariant representations of paired images and text \cite{saito2016demian}.
Modality-invariance was encouraged using an adversarial setup
	in which the discriminator was given one of the two
	representations or a sample drawn from the unit Gaussian.
The discriminator was tasked with determining
    which modality the input originated from or whether it was drawn from the unit Gaussian.
The encoders were trained through gradient reversal, as used previously in
	adversarial domain adaptation and generative adversarial networks \cite{saito2016demian,ganin2015unsupervised,ganin2016domain,tzeng2017adversarial,goodfellow14}.
Kashyap \cite{kashyap} also applied Harwath et al.'s \cite{harwath16} approach
    to the MNIST and TIDIGITs dataset, focusing
    primarily on using the embeddings for cross-modality transfer learning.
Our work focuses more on the embeddings themselves, and methods to promote modality-invariance.
Hsu et al.~\cite{hsu2017vae} designed a convolutional variational autoencoder (CVAE) for
    log mel-filterbanks of speech drawn from the TIMIT dataset.
In our work, we use the same convolutional network architecture for our audio encoder.
    
Our network architecture and loss function is based on Harwath et al., but instead
	of deterministically mapping inputs to embeddings, we map inputs to the parameterization of
    a diagonal Gaussian, and sample embeddings from it.
In addition, we add a regularization term for the posterior distributions.
In this regard, our method takes a similar approach to achieving modality-invariance as K. Saito et al.
	insofar as we both drive the distribution of embeddings to have minimal deviation from
	a unit Gaussian prior distribution of embeddings \cite{saito2016demian},
  but we found that our encoders can deceive
    a discriminator without using gradient reversal.
In addition, we believe the problem of modality-invariant embeddings using speech as
	one of the modalities has yet to be explored, so our research makes a novel contribution in this area.
    

\section{Methods}

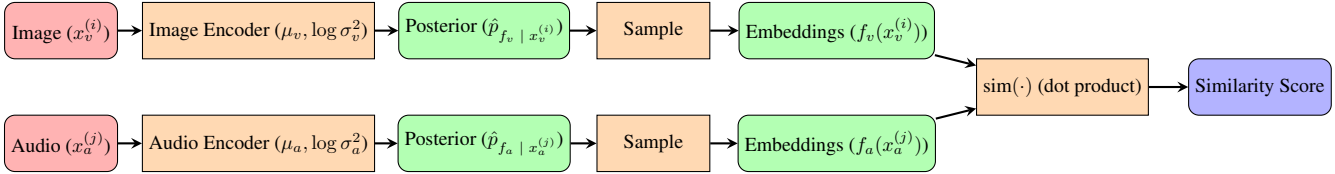
\begin{figure*}
	\centering
  \begin{tikzpicture}[node distance=2cm, scale=0.75, every node/.style={scale=0.75}]
    \node (inpi) [input] {Image ($\inp{\img}{i}$)};
    \node (inpa) [input, below of=inpi] {Audio ($\inp{\aud}{j}$)};
    \node (image) [hidden, right of=inpi, xshift=1.5cm] {Image Encoder ($\mu_\img, \log \sigma_\img^2$)};
    \node (audio) [hidden, right of=inpa, xshift=1.5cm] {Audio Encoder ($\mu_\aud, \log \sigma_\aud^2$)};
    \node (paramsi) [latent, right of=image, xshift=2cm] {Posterior ($\post{\img}{i}$)};
    \node (paramsa) [latent, right of=audio, xshift=2cm] {Posterior ($\post{\aud}{j}$)};
    \node (samplei) [hidden, right of=paramsi, xshift=1cm] {Sample};
    \node (samplea) [hidden, right of=paramsa, xshift=1cm] {Sample};
    \node (embeddingi) [latent, right of=samplei, xshift=1.25cm] {Embeddings ($\enc{\img}{i}$)};
    \node (embeddinga) [latent, right of=samplea, xshift=1.25cm] {Embeddings ($\enc{\aud}{j}$)};

    \node (dot) [hidden, right of=embeddinga, xshift=2cm, yshift=1cm] {$\text{sim}(\cdot)$ (dot product)};
    \node (similarity) [output, right of=dot, xshift=1.5cm] {Similarity Score};

    \draw [arrow] (inpi) -- (image);
    \draw [arrow] (inpa) -- (audio);
    \draw [arrow] (image) -- (paramsi);
    \draw [arrow] (audio) -- (paramsa);
    \draw [arrow] (paramsi) -- (samplei);
    \draw [arrow] (paramsa) -- (samplea);
    \draw [arrow] (samplei) -- (embeddingi);
    \draw [arrow] (samplea) -- (embeddinga);
    \draw [arrow] (embeddingi) -- (dot);
    \draw [arrow] (embeddinga) -- (dot);
    \draw [arrow] (dot) -- (similarity);
\end{tikzpicture}
  \vspace{0.4cm}
  \caption{The high-level model structure}
  \label{fig:architecture}
\end{figure*}

We first formalize the problem.
Given a set of co-occurring images and captions,
	$(\inp{\img}{i}, \inp{\aud}{i}), i=1...N$ where
    $\inp{\img}{i} \in \imgspace$ (image space) and $\inp{\aud}{i} \in \capspace$ (audio caption space),
	functions $f_\img \in \mathcal{F}_\img : \imgspace \mapsto \mathbb{R}^\repdim$
    and $f_\aud \in \mathcal{F}_\aud : \capspace \mapsto \mathbb{R}^\repdim$ are
    chosen to optimize some objective that promotes the encoding of semantic
    information contained in the inputs $x_\img^{(i)}$ and $x_\aud^{(i)}$
    into $\enc{\img}{i}$ and $\enc{\aud}{i}$, respectively.
For example, if $\inp{v}{i}$
	is a picture of a handwritten ``7'' and
	$\inp{\aud}{i}$ is an audio recording of someone saying ``seven'', 
    $\enc{\img}{i}$ and $\enc{\aud}{i}$ should be considered highly
    semantically related by some similarity metric.
As in \cite{harwath16}, we aim to increase the margin between the
    similarity of representations of co-occurring inputs and the
    similarity of representations of non-co-occurring inputs.
The {\it similarity loss} function is given in Equation~\ref{eq:lsim}.

In contrast to \cite{harwath16}, our encoders, $f_\img$ and $f_\aud$, are
	non-deterministic.
We learn the deterministic functions $\mu_\img : \imgspace \mapsto \mathbb{R}^\repdim$ and
	$\log \sigma^2_\img : \imgspace \mapsto \mathbb{R}^\repdim$.
Then we use $\mu_\img(\inp{\img}{i})$ and $\log \sigma^2_\img(\inp{\img}{i})$ to parameterize a
	diagonal Gaussian representing the posterior distribution over embeddings:
\begin{equation}
\label{eq:posterior}
\post{\img}{i} := \posterior{\img}{i}
\end{equation}
Embeddings are then sampled from the posterior:
\begin{equation}
\enc{\img}{i} \sim \post{\img}{i}
\end{equation}
and likewise for $f_\aud$.
We illustrate this process in Figure~\ref{fig:architecture}.

In addition to $\mathcal{L}_\text{sim}$, defined in Equation~\ref{eq:lsim},
	we average the KL divergence of the predicted posteriors over
    embeddings from
    the prior over embeddings (the unit Gaussian) as a regularization term
    we call {\it information gain (IG) loss}:
\begin{equation}
\label{eq:lig}
\begin{split}
\igloss = \frac{1}{N} \sum_i^N &KL(\post{\img}{i}\ ||\ \mathcal{N}(0,I_z)) \\
+ &KL(\post{\aud}{i}\ ||\ \mathcal{N}(0,I_z))
\end{split}
\end{equation}

Our total loss function is then:
\begin{equation}
\label{eq:loss}
\mathcal{L} = \simloss + \lamig \igloss +  \lamwd \wdloss
\end{equation}
where $\wdloss$ is the sum of all 
	Frobenius norms of weight matrices and convolutional kernels, and $\lamig$ and $\lamwd$
	are tunable hyperparameters.

\section{Datasets}
\label{sec:datasets}

For images, we used the MNIST dataset of handwritten digits \cite{mnist}.
The dataset contains 60K training images and 10K test images.
The images are 28x28 8-bit grayscale images, and we preprocess each image to have
pixel values between 0 and 1.
For audio, we used the TIDIGITS dataset of spoken utterances sampled at 20 KHz \cite{tidigits}.
We only used digit strings containing a single number, and we used utterances from men, women,
and children.
After filtering out utterances which contain more than one number, we have
    6,456 training utterances, 1,076 test utterances, and 1,076 validation utterances.
Using the Kaldi speech recognition toolkit \cite{kaldi},
    we generated 80 dimensional log mel-filterbank features with a 25ms window size and a 10ms frame shift,
    multiplied by a Povey window.
To create inputs of the same size, we pad or crop each spectrogram to
	100 frames (i.e., one second of speech) which is one frame longer than the mean frame length of
    the available utterances.
We preprocessed each filterbank to have zero mean and unit variance.
Longer utterances were center cropped.  Shorter utterances were zero padded at the end after adjusting
	the filterbank to have zero mean.
For TIDIGITS, we also combined the utterances labeled ``oh'' and ``zero'' into one class for the purpose
    of labeling clusters in our analysis\footnote{Training does not depend
    on explicit class labels except insofar as pairing audio and image inputs based on their ground truth digit labels.}.

\section{Experiments}
\label{sec:experiments}

We used convolutional neural networks to predict the parameterizations of
	$\post{\img}{i}$ and $\post{\aud}{i}$ (Equation~\ref{eq:posterior}).   
We trained the networks to minimize Equation~\ref{eq:loss} for the MNIST and TIDIGITS datasets
	described in Section~\ref{sec:datasets}.
We compared the embedding spaces produced when $\lamig = 0$ and when $\lamig > 0$ to gauge the
	effect of regularizing information gain in the posterior.

We set the embedding dimension to be $\repdim = 128$, which is consistent with the latent
	embedding
	dimensionality used by \cite{hsu2017vae} for their variational autoencoder for 58 phones.
We did not explore other values of $\repdim$.
The encoders for both images and audio are convolutional networks
	which produce the parameterization (the mean and log variance vectors)
    of the posterior distribution over embeddings.
The audio encoder uses the same architecture
    as the encoder portion of Hsu et al.'s variational autoencoder
    for 80 dimensional log mel-filterbank speech \cite{hsu2017vae}.

The image encoder is also convolutional, taking the following form:
\begin{enumerate}[topsep=0pt,itemsep=-1ex,partopsep=1ex,parsep=1ex]
\item $3 \times 3$ conv., 64 filters, same padding
\item $3 \times 3$ conv., $2 \times 2$ strides, 128 filters, same padding
\item $3 \times 3$ conv., $2 \times 2$ strides, 256 filters, same padding
\item $512$ unit fully connected
\item $256$ unit linear output ($128$ for $\mu_\img$, $128$ for $\log \sigma_\img^2$)
\end{enumerate}
ReLu activations were used for each layer except the final linear layer.
A weight decay ($\lambda_\text{WD}$) of $10^{-6}$ is used for all convolutional and fully connected layers.
The initial learning rate was $10^{-5}$ which was decayed by a factor of 0.9 every 10 epochs.
The Adam optimization algorithm was used with $\beta_1 = 0.95$, $\beta_2 = 0.999$, and $\epsilon = 10^{-8}$.
128 distinct image-audio pairs were used for each batch.  After processing each image or audio input through the respective encoder to produce a posterior distribution, 16 embeddings were sampled per input.\footnote{Positive image-audio embedding pairings were established by matching corresponding sampled embeddings for each input.}  This produced a total of 2,048 image-audio embedding pairs in each batch.  

Negative sampling was performed by selecting one of the other 2,047 sample pairs in the batch.
While it would at first seem reasonable to disallow negative samples for a training pair to be drawn from the same underlying digit class,
	such a mechanism implies a ground truth digit labelling of all examples within a batch.
In other words, the knowledge of which negative example pairs \textit{not} to sample is equivalent to the network possessing
	an oracle that knows which audio/visual sample pairs within a batch were drawn from the same underlying digit class.
This oracle would allow the network to trivially recover the ground truth digit labelling of all examples within a batch.
In an effort to avoid this, we allow negative samples to be chosen from any digit class regardless of the
	initial example's digit class.
Empirically, we found that the weight of the positive examples can easily overcome the ``contradictory'' signals
	introduced by this sampling scheme, allowing the model to produce a semantically rich embedding space.

The model was trained for 100 epochs.
An epoch was defined as the number of batches required to cover all training examples in the
	larger of the two datasets (MNIST) exactly once.
Training required about 35 minutes on an NVIDIA TitanX GPU.

\section{Results and Analysis}
\label{sec:results}

To analyze the learned semantic space, we sampled embeddings for inputs from the unseen test set,
	sampling 16 samples per input point.
We ran K-means clustering with $k=10$ and calculated the cluster purity of the resulting clusters,
    defined as:
\begin{equation}
\label{eq:purity}
\frac{1}{N} \sum_{i=1}^k \max_{j=1...k} \left( \left| c_i \cap y_j \right| \right)
\end{equation}
where $c_i$ is the set of all points in cluster $i$ and $y_j$ is the set of all points
    of class $j$ (their ground truth digit label).
This metric represents the accuracy of a classifier which classifies a point, $x$, according
    to the majority class of the cluster whose mean is closest to $x$ using euclidean distance.

We then used a subset of 2,152 sample embeddings (1,076 from images, 1,076 from audio) and performed
    a classification task to predict the original input point's modality from the embeddings
    using an SVM with a Gaussian RBF kernel.
1600 examples were used for the training set and the remaining were used for the test set.
We used 3-fold validation to select a $C$ value for the SVM.
Comparing the modality classification test accuracy to the prior on modality ($\frac{1}{2}$) allows us to gauge the extent to
    which the embeddings are modality-invariant.
Perfectly modality-invariant embeddings would result in a modality classification test accuracy of $\frac{1}{2}$.

\begin{figure}[!h]
 \centering
  \includegraphics[width=1.0\linewidth]{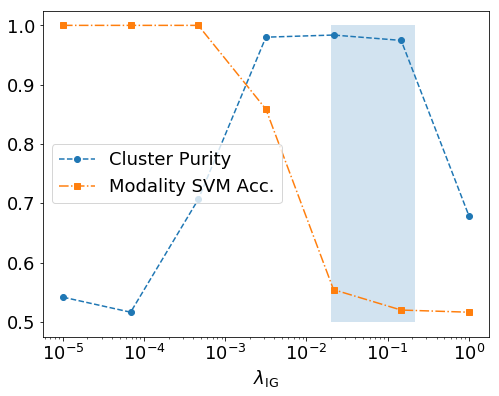}
  \caption{Effects of tuning the weight, $\lambda_\mathrm{IG}$, of the KL divergence regularization term. The shaded region is considered ideal: modality classification is nearly random and cluster purity reaches its peak.}
  \label{fig:tuning}
\end{figure}

We evaluated the effect of $\lamig$ on the cluster purity and modality invariance of the
	embeddings learned by our model.
Results from using the modality classifier and cluster purity analysis
	are shown in Figure~\ref{fig:tuning} and Table~\ref{table:tuning}.

\begin{figure}[!h]
     \centering
     \vskip 0cm
      \includegraphics[width=\linewidth]{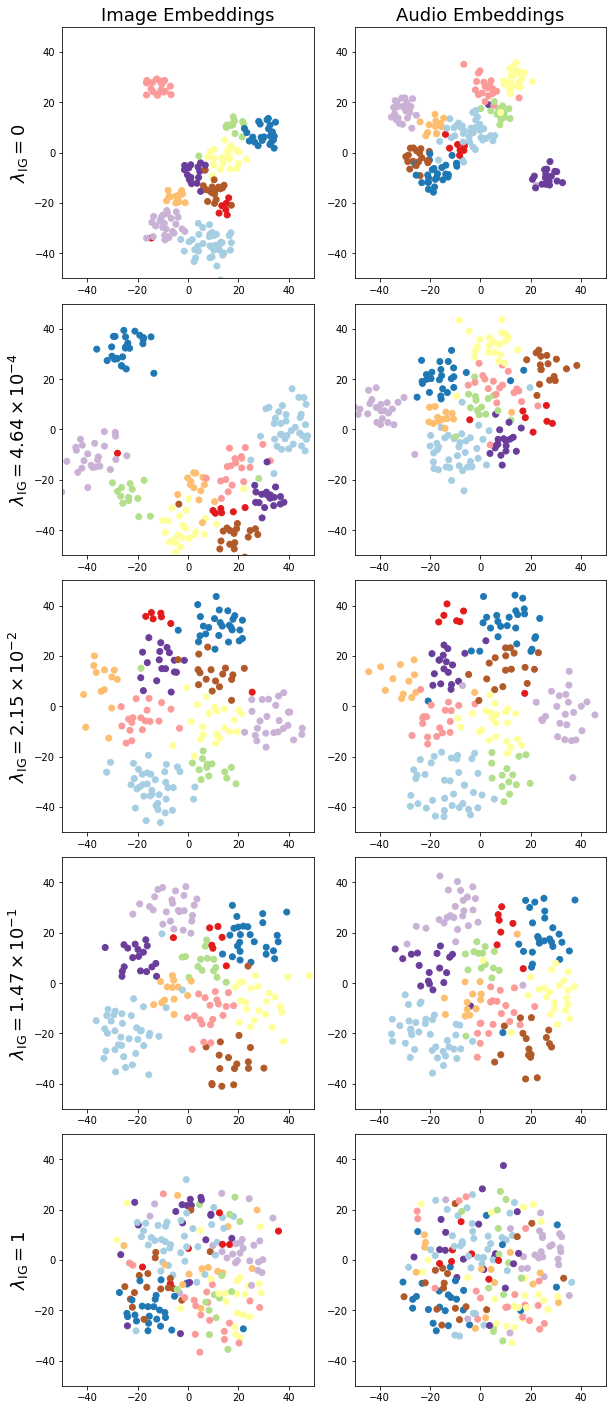}
     \vskip 8pt
      \caption{2 dimensional t-SNE projections of 128 dimensional embeddings produced from using various weights, $\lambda_\mathrm{IG}$, for the KL divergence regularization term.}
      \vskip 13pt
     \label{fig:tsne}
\end{figure}

\begin{table}[!h]
\centering
\begin{tabular}{ccc}
$\lamig$ & Cluster Purity & Modality SVM Acc. \\ \hline \hline 
0.00e+00 & 0.525 & 1.000 \\ \hline 
1.00e-05 & 0.542 & 1.000 \\ \hline 
6.81e-05 & 0.516 & 1.000 \\ \hline 
4.64e-04 & 0.707 & 1.000 \\ \hline 
3.16e-03 & 0.980 & 0.859 \\ \hline 
2.15e-02 & \textbf{0.984} & 0.554 \\ \hline 
1.47e-01 & 0.975 & 0.520 \\ \hline 
1.00e+00 & 0.679 & \textbf{0.516} \\ \hline 
\end{tabular}
\vskip 6pt
\caption{Cluster purities and modality
	classification accuracies for various values of $\lamig$}
\label{table:tuning}
\end{table}

In addition, we used 200 samples per modality to compute a two dimensional t-SNE projection\footnote{t-SNE
    was selected over PCA for its ability to show relative pairwise distances \cite{tsne}}
    of the embeddings produced by each hyperparameter setting.
We plotted these samples in Figure~\ref{fig:tsne} and colored them according to class label.
For both cells in a row, the same TSNE model was used, so the embeddings for both modalities were
	projected into the same two-dimensional space.
    
The additional $\igloss$ term resulted in
    greater cluster purity, as shown in Figure~\ref{fig:tuning}.
The lower cluster purity for $\simloss$ alone ($\lamig = 0$) is visually evident
	in the first row of Figure~\ref{fig:tsne}:
    though there are clear semantic clusterings of samples from the same digit, there
	are typically two clusters per digit---one for images and one for audio.
One possible explanation for why the cluster purity is low ($0.525$) for $\lamig = 0$
	is that when K-Means is performed
	with $k = 10$, $k$ is about half the number of digit clusters in the embedding space
    (one for each digit-modality pair), resulting in K-Means clusters with members 
    nearly evenly split between two digits.
This finding shows that while using $\simloss$ alone, embeddings originating from the same modality may still be significantly
    closer together than embeddings of different modalities, regardless of the similarity
    of semantic content.
The 100\% accuracy of the SVM in predicting the modality of embeddings when $\lamig = 0$,
    as shown in Figure~\ref{fig:tuning}(a), further supports the finding that the embedding space produced from using
    $\simloss$ alone is not modality invariant.
This trend is not conducive to modality
    invariance of the embeddings.
    
In contrast, the embeddings produced when using $\igloss = 2.15 \cdot 10^{-2}$
	for training were only able to be
    classified by an SVM with 55.4\% accuracy, as shown in
    Table~\ref{table:tuning}.
Although this metric is not the ideal 50\% accuracy of truly
	modality-invariant embeddings, the embedding space produced
    using $\igloss$ is much closer to being modality-invariant than
    the space produced by $\simloss$ alone.
    
Figure~\ref{fig:tuning} shows that minimizing the divergence of the posterior over embeddings from the prior improves
	modality invariance.
This could be due to the fact that the KL divergence represents the amount of information about an embedding conveyed by an input,
	and by limiting the amount
    of information, we force the encoders to filter out information.
This is the same reason why variational autoencoders exhibit de-noising behavior \cite{vae}.
Since semantic information is important for minimizing $\simloss$,
	modality information tends to be filtered out before
    semantic information.
For our applications, Figure~\ref{fig:tuning}(ii) shows
	that $\lamig \approx 2.15 \cdot 10^{-2}$
	is the empirically observed ideal cutoff point
    at which increasing $\lamig$ to further limit
    the total information conveyed in the posterior
    begins to also to overly restrict the semantic information conveyed, resulting
    in a drop in cluster purity.
Figure~\ref{fig:tsne} shows this trend qualitatively. Row 1 shows sampled embeddings
	resulting from an under-regularized model; row 3, well regularized;
    and row 5, over regularized.

\section{Conclusions}

In this work, our goal was to learn a joint modality-invariant semantic embedding space
	for speech and images in an unsupervised manner.
We focused on spoken utterances and images of handwritten digits.
We found that by sampling encodings rather than predicting them directly, and by regularizing the posterior distribution over embeddings,
	we were able to learn a more modality-invariant semantic embedding space.
From an adversarial perspective, we were able to
	deceive an adversarial discriminator (the
    modality-classifying SVM) without the use of gradient reversal
    or any adversarial setup during training.
This leads us to suspect $\igloss$ may be a
	useful regularization term in
	other generative adversarial approaches to learning
    domain or modality invariant embeddings.
    
Further research could be done to attempt to
	combine the techniques used by VAEs and GANs.
One potential direction in the vein of multimodal learning
	of a semantic space is to replace the dot product similarity
    with a symmetric divergence of the variational distributions of matched and mismatched inputs.
This would allow for a more probabilistically theoretical formulation of the loss function which could have more general
    implications for other areas of research.

\subsection*{Acknowledgements} The authors would like to thank Wei-Ning Hsu for his help with the audio encoder architecture, and advice on variational auto-encoders.

\clearpage

\bibliographystyle{IEEEbib}
\bibliography{mybib}

\end{document}